\documentclass[letterpaper, 10 pt, conference]{ieeeconf}  

\IEEEoverridecommandlockouts                              
\usepackage[l2tabu,orthodox]{nag}

\usepackage[
    backend=bibtex8,
    style=ieee,
    sorting=none,
    natbib=true,
    doi=false,
    isbn=false,
    url=false,
    eprint=false,
    maxcitenames=1,
    mincitenames=1
]{biblatex}

\usepackage[pdftex,colorlinks]{hyperref}

\usepackage[printonlyused]{acronym}

\usepackage{siunitx}
\usepackage[all]{nowidow}

\usepackage[dvipsnames]{xcolor}

\usepackage{lipsum}

\usepackage{xspace} 
\newcommand{\ie}{i.e.,\xspace{}}

\usepackage[pdftex]{graphicx}

\usepackage{epstopdf}

\usepackage{import}

\graphicspath{{./latexGoodPractices/}}

\usepackage{booktabs}

\usepackage{tabularx}
\usepackage{multirow, multicol}

\usepackage{amssymb,amsfonts,amsmath,amscd}

\usepackage{bm}

\newcommand{\bbm}{\begin{bmatrix}}
\newcommand{\ebm}{\end{bmatrix}}

\usepackage{physics}  
\usepackage{amsmath} 
\usepackage[normalem]{ulem}
\usepackage{siunitx}
\usepackage{stfloats}   

\makeatletter
\let\oldtheequation\theequation
\renewcommand\tagform@[1]{\maketag@@@{\ignorespaces#1\unskip\@@italiccorr}}
\renewcommand\theequation{(\oldtheequation)}
\makeatother

\title{\LARGE \bf 
    TACOS: Terrain-Aware COntact Supervision for \\ 
    Visual Footstep Planning
}

\author{Jean-Michel Fortin$^{1,*}$, William Larrivée-Hardy$^{1}$, Gabriel Jeanson$^{1}$, Olivier Gamache$^{1}$, \\ Daniele De Martini$^{2}$, François Pomerleau$^{1}$, Philippe Giguère$^{1}$
\thanks{$^{1}$Northern Robotics Laboratory, Université Laval, Canada}%
\thanks{$^{2}$Mobile Robotics Group, University of Oxford, United Kingdom}
\thanks{$^{*}$Corresponding author: \texttt{jean-michel.fortin.8@ulaval.ca}}
}%

\usepackage[switch]{lineno}
\usepackage[font=small]{caption}
\usepackage{subcaption}

\acrodef{SLAM}{Simultaneous Localization And Mapping}
\acrodef{UGV}{Uncrewed Ground Vehicles}
\acrodef{UAV}{Uncrewed Aerial Vehicle}
\acrodef{FPV}{First-Person View}
\acrodef{LAGR}{Learning Applied to Ground Vehicles}
\acrodef{MCR}{Motor Currents versus Rate of turn}
\acrodef{ICR}{Instantaneous Centre of Rotation}
\acrodef{IMU}{Inertial Measurement Unit}
\acrodef{NMHE}{Nonlinear Moving Horizon Estimator}
\acrodef{CNN}{Convolutional Neural Networks}
\acrodef{TnR}{Teach-and-Repeat}
\acrodef{fps}{frames per second}
\acrodef{GNSS}{Global Navigation Satellite System}
\acrodef{PTP}{Precision Time Protocol}
\acrodef{BEV}{Bird's-Eye View}
\acrodef{PoV}{Point-of-View}
\acrodef{FoV}{Field-of-View}
\acrodef{PSD}{Power Spectral Density}
\acrodef{GSD}{Ground Sampling Distance}
\acrodef{RTK}{Real-Time Kinematic}
\acrodef{AUC}{Area Under the Curve}
\acrodef{POI}{Points of Interest}
\acrodef{MLP}{Multi-Layer Perceptron}
\acrodef{MPC}{Model Predictive Control}
\acrodef{FFT}{Fast Fourier Transform}
\acrodef{RMSE}{Root Mean Squared Error}
\acrodef{PCA}{Principal Components Analysis}
\acrodef{GRF}{Ground Reaction Force}
\acrodef{EKF}{Extended Kalman Filter}
\acrodef{UKF}{Unscented Kalman Filter}
\acrodef{MAE}{Mean Absolute Error}
\acrodef{FK}{forward kinematics}
\acrodef{HMM}{Hidden Markov Model}
\acrodef{WVN}{Wild Visual Navigation}
\acrodef{GRS}{Ground Reaction Score}
\acrodef{DPT}{Dense Prediction Transformer}
\acrodef{PR}{precision-recall}
\acrodef{AP}{Average Precision}
\acrodef{VUS}{Value Under the Surface}
\acrodef{FiLM}{Feature-wise Linear Modulation}
\acrodef{TACOS}{Terrain-Aware COntact Supervision}
\acrodef{GCA}{Gated Cross-Attention}
\acrodef{RL}{reinforcement learning}
\acrodef{EMA}{exponential moving average}

\newcommand{\circled}[1]{\raisebox{.5pt}{\textcircled{\raisebox{-.3pt}{\scriptsize #1}}}}

\begin{document}
\maketitle
\thispagestyle{empty}
\pagestyle{empty}

\begin{abstract}

Humans navigate their environment based on their vision system, anticipating how the ground will respond before committing to a step.
Self-supervised learning could provide legged robots the same ability, but the supervisory signal is typically chosen on intuition, and its effect on what the model learns is rarely examined.
We present \ac{TACOS}, which proposes a meaningful and consistent footstep instability score, based on the residual of a factor graph in legged state estimation. 
It projects the continuous score into preceding images to supervise a dense visual predictor.
We propose two variants, a \textit{hard} score that exploits hindsight and a causal \textit{soft} score available immediately.
Both metrics are compared against state-of-the-art alternatives in two independent evaluations that share no data or labels.
On $1716$ manually annotated contacts, both variants outperform every evaluated alternative at detecting instability and ranking its severity, improving pooled severity ranking by \SI{28}{\%} over the strongest baseline.
On three outdoor test runs, our metric also trains the most accurate visual predictor, improving rank correlation by \SI{58}{\%} over the strongest alternative, and by a further \SI{25}{\%} when supplemented with onboard context.
In a real-world experiment over a \SI{50}{m} forest trajectory, the trained predictor handles non-geometric terrain properties such as mud and vegetation, while stepping around geometric obstacles like fallen branches. 
Code is available for reproducibility at \url{https://github.com/norlab-ulaval/TACOS}.

\end{abstract}


\acresetall 

\section{Introduction}
\label{sec:introduction}



\begin{figure}[t]
	\centering
	\includegraphics[width=0.48\textwidth]{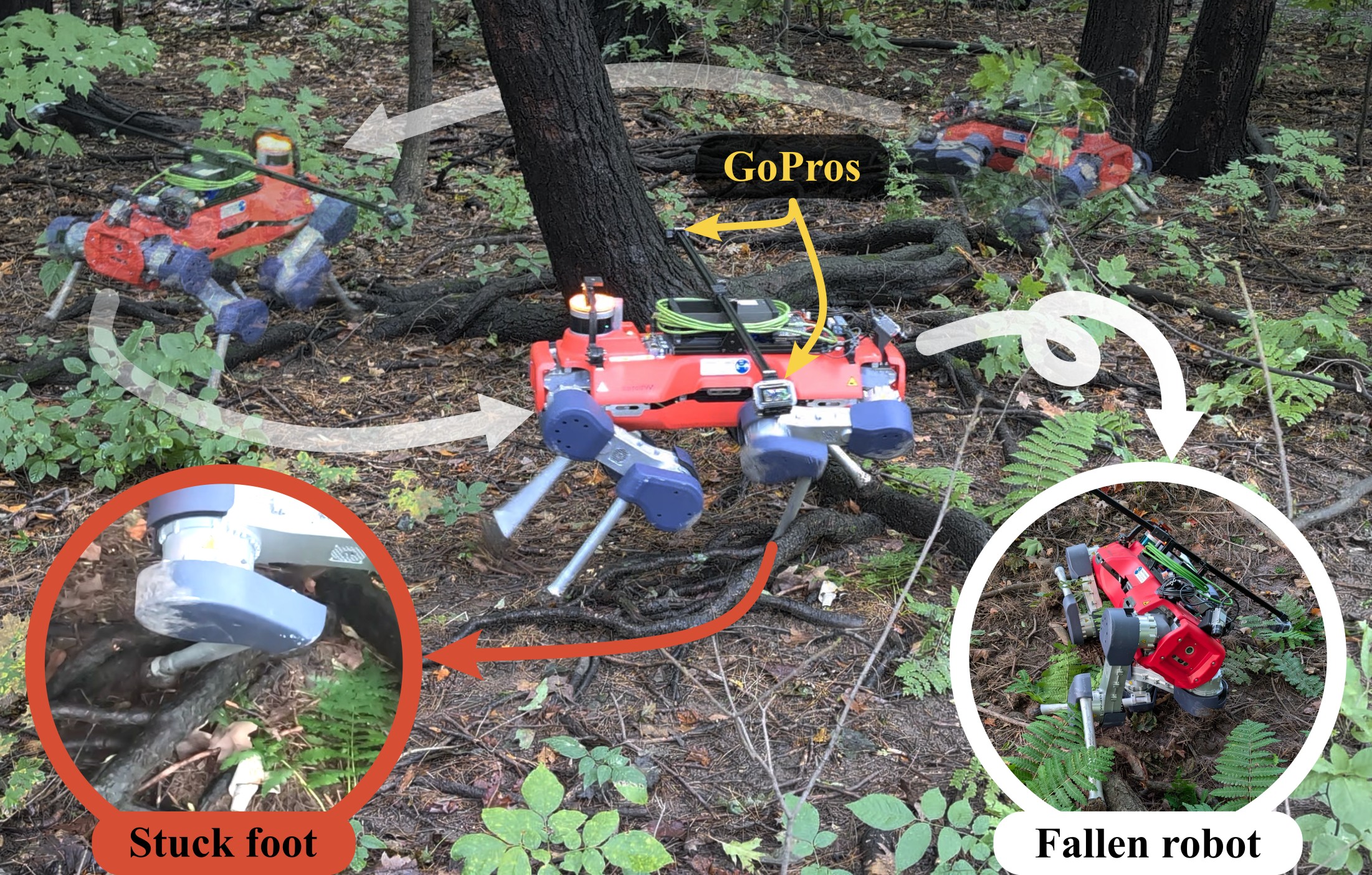}
	\caption{
    Quadruped robot traversing unstructured forest terrain, where roots, branches, and leaves create candidate footholds whose supportiveness cannot be inferred from geometry alone.
    Close-ups show, on the right, the GoPro view of the foot getting stuck, wedged among exposed roots, and on the left, the resulting robot fall.
    } 
	\label{fig:intro}
    \vspace{-10pt}
        
\end{figure}

Quadruped robots demonstrate remarkable agility in complex environments~\citep{miki2022learning, qin2026comprehensive}, enabling applications in search and rescue, forest monitoring, and planetary exploration.
Exploiting this mobility in unstructured natural terrain requires selecting reliable footholds, whose behavior during contact cannot be determined from geometry alone~\citep{wellhausen2019should}.
A seemingly suitable surface may offer insufficient friction or deform under load, while vegetation can conceal exposed roots that trap a foot and cause a fall, as illustrated in~\autoref{fig:intro}.
Avoiding such failures requires anticipating how the foot will interact with the terrain before committing to a step~\citep{fahmi2022vital}.
Humans use visual cues to plan future steps, but such anticipation remains an open challenge for legged robots~\citep{frey2026advances}.

Most footstep planners rely on geometric representations~\citep{jenelten2020perceptive}, while \ac{RL}, effective for proprioceptive locomotion, remains sample-inefficient on high-dimensional visual inputs~\citep{frey2026advances}.
Self-supervised learning offers an alternative by leveraging the robot's own interactions with the terrain as supervision. 
However, the resulting visual representation depends directly on the quality of the supervisory signal.
Obtaining scalable, physically meaningful labels for foothold instability is particularly challenging, since manual annotation is subjective~\citep{schreiber2024w} and terrain traversability is inherently robot- and task-dependent~\citep{frey2026advances}.

Existing self-supervised traversability methods commonly use body-level quantities, such as tracking error~\citep{mattamala2025wild} or current consumption~\citep{carvalho2026self}, which describe the robot's overall response rather than individual footholds.
Others instead used footstep-level supervision~\citep{wellhausen2019should}, gaining spatial precision but relying on fragile foot force-torque sensors rarely available on commercial platforms.
Legged state estimation literature proposed methods to detect unreliable contacts from proprioception~\cite{yoon2023invariant,girgin2026ocelot}, but uses them to correct odometry reactively rather than to anticipate instability from exteroception.

In this work, we propose \ac{TACOS}, a self-supervised framework for predicting foothold instability from vision in unstructured terrain. 
Our key insight is that the rigid-contact assumptions underlying legged state estimation provide a natural measure of foothold instability. 
When a foot slips, sinks, or loses support, the resulting inconsistency appears in the residual of the corresponding contact constraint. 
We aggregate these residuals over each stance into a continuous per-footstep metric, in a \textit{hard} variant that exploits the full estimation window and a \textit{soft} variant available online.
Projecting the observed footholds into preceding images yields sparse labels that supervise a dense visual predictor.
We evaluate the metric against manually annotated contact instability, assess its suitability for visual learning on unseen trajectories, and demonstrate its use in closed-loop footstep planning.

In short, our contributions are :
\begin{itemize}
    \item A novel continuous per-footstep instability score derived from factor graph residuals, in a \textit{hard} variant used as our reference label and a \textit{soft} variant suitable for online use, both validated against manually annotated unstable contacts;
    \item An analysis of the influence of the supervisory metric and of context conditioning on visual instability learning from sparse footstep labels;
    \item A closed-loop demonstration of the learned visual instability predictor for footstep planning in unstructured forest terrain.
\end{itemize}

\section{Related Work}
\label{sec:related_work}


\subsection{Legged State Estimation}

Legged state estimation commonly exploits the assumption that a stance foot remains stationary relative to the terrain. 
Early approaches combined inertial measurements with leg kinematics and contact information to constrain the robot state and reduce inertial drift~\citep{bloesch2013state}. 
This principle was subsequently incorporated into invariant filtering~\citep{hartley2020contact} and smoothing formulations~\citep{wisth2019robust,yoon2023invariant}, where individual stance contacts constrain the estimated robot motion relative to foothold locations.
The reliability of these constraints depends on the validity of the assumed contact model. 
Slip, compliance, and terrain deformation can introduce disagreement between observed kinematics and the stationary-foot assumption. 
Contact-aware estimators therefore identify unreliable measurements and adjust their contribution to the state estimate~\citep{camurri2017probabilistic,yoon2023invariant,girgin2026ocelot}. 
Beyond their effect on localization, these inconsistencies provide observations of the underlying foot-terrain interaction, on which our instability score is based.

\subsection{Foothold Instability Estimation}

Contact instability is not directly observable and must be inferred from the robot's response during stance. 
Although much of the literature focuses on slip detection, natural terrain also introduces sinkage and loss of support, motivating a broader assessment of foothold reliability.
Force-based approaches characterize contact through measured or estimated \mbox{\acp{GRF}}. 
Foothold robustness has been defined as the force a contact can sustain and predicted through haptic exploration~\citep{hoepflinger2013unsupervised}. 
Similar work has used internal force sensing to estimate reliable-contact probabilities for leg odometry~\citep{camurri2017probabilistic}. 
For terrain characterization, \citet{wellhausen2019should} derived a \ac{GRS} through dimensionality reduction of force-torque measurements from sensorized feet. 
Estimating \mbox{\acp{GRF}} from actuator torques removes the need for foot sensors, but introduces sensitivity to joint friction, backlash, and model errors~\citep{hwangbo2016probabilistic}.
\citet{jenelten2019dynamic} instead modeled the uncertainty of contact and slip with probabilistic state machines, whose slip probability expresses confidence in a slipping state, not its severity.
More recently, OCELOT~\citep{girgin2026ocelot} combined force-based contact detection with a kinematic stationarity test.
Their continuous metric indicates whether a foot is both loaded and stationary, connecting force sensing with motion consistency to quantify contact reliability for leg odometry.

Foothold instability can also be evaluated without force sensing.
Early approaches compared each stance foot's velocity with the median across stance feet, requiring at least three simultaneous contacts~\citep{focchi2017slip}.
Comparing foot motion with its commanded trajectory removes this requirement and detects concurrent slips~\citep{nistico2022slip}, but assumes planned foot trajectories, which modern reinforcement learning controllers do not provide.
Other methods examine stance-foot motion through a robot state estimate.
Using a proprioceptive Kalman filter, \citet{yan2024slip} derived a slip-likelihood score from each stance foot's displacement relative to its touchdown location and its lateral velocity.
By retaining a history of robot states, the invariant smoother of \citet{yoon2023invariant} re-evaluates estimated foot velocities as new measurements become available, improving the identification of dynamic contacts.
These approaches use deviations from expected contact behavior to identify unreliable support.
Our method build on this principle by aggregating contact-model residuals over each stance into a continuous per-footstep instability score, providing localized supervision for visual instability prediction.

\subsection{Self-Supervised Terrain Learning}

Manual terrain annotation is expensive and difficult to scale across diverse natural environments~\citep{schreiber2024w}. 
Self-supervised approaches instead obtain supervision from the robot's physical experience. 
\ac{WVN} derives a traction metric from the discrepancy between commanded and measured body velocity~\citep{mattamala2025wild}, P-STEER uses the instantaneous current as an indicator of traversal effort~\citep{carvalho2026self}, and ProNav combines actuator and battery signals into a low-dimensional representation~\citep{elnoor2024pronav}.
Others instead learn a continuous embedding of the interaction, scoring terrain from operator preferences~\citep{karnan2023sterling} or by proximity to a reference terrain~\citep{hindel2026self} rather than from a physical quantity.

Self-supervision from robot experience is inherently sparse, since the robot probes the terrain only where it steps.
All the aforementioned methods expand these sparse labels to larger image regions through learned segmentation or visual-feature grouping~\citep{carvalho2026self,mattamala2025wild,hindel2026self}, or replace the egocentric image by footprint-sized patches~\citep{karnan2023sterling}.
\citet{wellhausen2019should} instead associates each per-footstep \ac{GRS} with its foothold location and projects it into preceding images, supervising only where feet physically touched the terrain.
These association strategies trade supervision coverage for spatial fidelity.
Extending labels to visually similar regions assumes comparable physical support, which may be unreliable in heterogeneous natural terrain~\citep{miki2022learning,mattamala2025wild}.
Sparse annotations have nonetheless been shown to support dense traversability prediction~\citep{schreiber2024w}, motivating supervision without densification.
We therefore retain supervision only at physically observed footholds~\citep{wellhausen2019should}, and examine how the choice of instability score affects the visual predictor's ability to anticipate contact instability.

\section{Methodology}
\label{sec:methodology}

\begin{figure*}[htbp]
    \centering
    \includegraphics[width=0.97\textwidth]{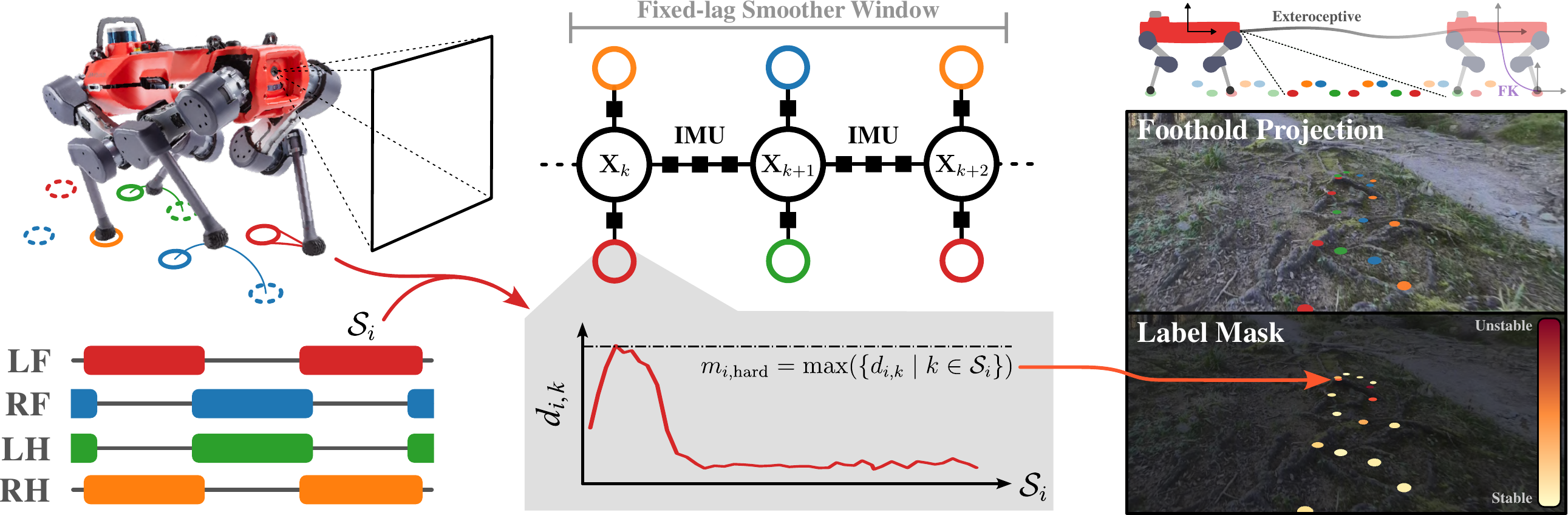}
    \caption{Offline label-generation pipeline. \textbf{(Left)} Stances are extracted from the onboard contact detector, where F and H denote front and hind, and L and R left and right. \textbf{(Center)} Contact residuals $d_{i,k}$ from a factor graph leg odometry are aggregated over each stance into the instability score $m_{i,\mathrm{hard}}$. \textbf{(Right)} Footstep placements are computed in the world frame using exteroceptive odometry and \acf{FK}, then projected into an image acquired before contact and rasterized into a sparse label mask.}
    \label{fig:metho_label_generation}
    \vspace{-10pt}
\end{figure*}

We formulate foothold instability estimation as a self-supervised learning problem in which the robot's interactions with the terrain provide supervision for visual perception. 
While walking, each leg alternates between stance, when its foot supports the robot, and swing, when it travels to the next foothold.
Each stance is treated as a local probe of the terrain, \ie{} measurements collected during the stance characterize the quality of the foot-terrain interaction.
Meanwhile, the touchdown location identifies the corresponding terrain region.

Our methodology consists of three main steps. 
First, we derive a per-footstep instability score from factor graph contact residuals.
Second, each measurement is associated with its foothold location and projected into synchronized camera images to generate sparse visual supervision. 
These two stages are performed offline, and together form the label-generation pipeline summarized in \autoref{fig:metho_label_generation}.
Finally, these sparse observations are used to train a dense visual predictor of foothold instability, as described in \autoref{sec:metho_instability_predictor}.

Throughout this section, $i$ indexes stance events and $k$ indexes proprioceptive robot-state measurements, with $\mathcal{S}_i$ denoting the set of sample indices belonging to stance $i$. 
Frames are denoted by calligraphic symbols, with $\mathcal{W}$, $\mathcal{B}$, and $\mathcal{C}$ representing the world, robot-base, and camera frames, respectively. 
We denote by ${}^{\mathcal A}_{\mathcal B}\mathbf{T} \in SE(3)$ the rigid transformation from frame $\mathcal{B}$ to frame $\mathcal{A}$, and by ${}^{\mathcal A}\mathbf{p}$ a point expressed in frame $\mathcal{A}$.

\subsection{Contact Detection}

Since foothold instability is computed only over samples assigned to a stance, contact-detection errors directly degrade label quality. 
We initialize stance intervals from the contact states provided on the ANYmal robot, which are estimated from foot kinematics and \acp{GRF}~\cite{hwangbo2016probabilistic}. 
Each detected contact interval defines a stance $i$, with samples $\mathcal{S}_i$ over which the instability score is computed, shown on the left of \autoref{fig:metho_label_generation}.
The raw contact signal occasionally contains short spurious transitions, which we remove with a morphological debounce filter. An opening step discards contacts shorter than \SI{100}{ms}, and a subsequent closing step bridges gaps shorter than \SI{25}{ms}.
We also observed that some detected contacts extend beyond the physical end of stance and into foot liftoff.
Since these samples no longer characterize the foot-terrain interaction, we refine the end of each stance offline. Within a \SI{150}{ms} window preceding the detected liftoff, we search backward for the last sample whose \ac{GRF} exceeds \SI{50}{N} and truncate the stance there, discarding the unloaded tail.
These four thresholds were determined empirically by inspecting stable and unstable stance windows in the recorded data.
The resulting stance intervals are used for all subsequent computations.

\subsection{Residual-based Instability Score}
\label{sec:foothold_metric}


Given the detected stance intervals, we assess foothold instability through violations of the stationary-foot assumption underlying leg odometry.
Following \citet{dellaert2026four}, we formulate proprioceptive odometry as a fixed-lag factor graph combining \ac{IMU} preintegration~\citep{forster2015imu}, bias estimation, and per-leg contact factors, solved incrementally using GTSAM~\cite{gtsam}.
Each keyframe $k$ contains the floating-base state $\mathbf X_k=(\mathbf R_k,\mathbf p_k,\mathbf v_k)$ and \ac{IMU} bias $\mathbf b_k$, where $\mathbf R_k\in SO(3)$ maps base-frame vectors to the world frame, and $\mathbf p_k,\mathbf v_k\in\mathbb R^3$ denote the world-frame base position and velocity.
For stance $i$, the measured base-frame foot position ${}^{\mathcal B}\mathbf s_{i,k}$ is obtained from joint positions through \acf{FK}.

We consider two variants of the instability score, \textit{hard} and \textit{soft}, both using the residual
\begin{equation}
    \mathbf e_{i,k} = \mathbf R_k^\top(\mathbf f_{i,k}-\mathbf p_k) - {}^{\mathcal B}\mathbf s_{i,k},
    \label{eq:contact_residual}
\end{equation}
where $\mathbf f_{i,k}\in\mathbb R^3$ denotes the world-frame foothold position.
At touchdown, the foothold is initialized from the base pose and \ac{FK}, with a weak prior of covariance $\sigma_f^2\,\mathbf I_3$, where $\sigma_f=\SI{0.5}{m}$.
A Huber kernel with standard $\delta=1.345$ limits the influence of an unstable foothold on the estimated trajectory.
We normalize the residual using its Mahalanobis norm
\begin{equation}
    d_{i,k}
    =
    \sqrt{\mathbf e_{i,k}^{\top}
    \boldsymbol{\Sigma}_{\mathrm{ct}}^{-1}
    \mathbf e_{i,k}},
    \qquad
    \boldsymbol{\Sigma}_{\mathrm{ct}}
    =\sigma_{\mathrm{ct}}^2\,\mathbf I_3,
    \label{eq:contact_inconsistency}
\end{equation}
with $\sigma_{\mathrm{ct}}=\SI{8}{cm}$ selected empirically.
The isotropic normalization makes the score rotation-invariant.

In the \textit{hard} formulation, a single foothold state is shared across the stance, so $\mathbf f_{i,k}=\mathbf f_i$.
The residual therefore measures disagreement with a stationary foothold.
We read each residual only as its stance leaves the fixed-lag window, once later measurements have refined the states it spans, as shown in the center of \autoref{fig:metho_label_generation}.
We aggregate these residuals using their maximum over the stance interval, yielding
\begin{equation}
    m_{i,\mathrm{hard}}
    =
    \max\!\left(
    \{d_{i,k}\mid k\in\mathcal S_i\}
    \right).
    \label{eq:foothold_metric_hard}
\end{equation}

In the \textit{soft} formulation, the shared foothold state is replaced by per-sample states linked by a random walk:
\begin{equation}
    \mathbf f_{i,k+1}
    =
    \mathbf f_{i,k}+\mathbf w_{i,k},
    \qquad
    \mathbf w_{i,k}
    \sim\mathcal N
    \left(\mathbf 0,\mathbf Q_f\Delta t_k\right),
    \label{eq:foothold_random_walk}
\end{equation}
where $\Delta t_k$ is the sampling interval and
$\mathbf Q_f=\sigma_{\mathrm{rw}}^2\,\mathbf I_3$, with
$\sigma_{\mathrm{rw}}^2=\SI{0.04}{m^2/s}$, also selected empirically.
Only the initial foothold state receives the touchdown prior.
Allowing gradual foothold motion absorbs part of the discrepancy into the foothold state rather than the residual.

This formulation also supports causal evaluation by computing each residual immediately after the smoother update incorporating its sample, enabling online feedback on foothold stability.
We reduce sensitivity to isolated peaks by aggregating the residuals over the completed stance using its 95th percentile, yielding
\begin{equation}
    m_{i,\mathrm{soft}}
    =
    P_{95}\!\left(
    \{d_{i,k}\mid k\in\mathcal S_i\}
    \right).
    \label{eq:foothold_metric}
\end{equation}

The \textit{hard} metric is therefore the reference formulation, and the one we use to supervise the visual predictor, while the \textit{soft} one is an instantaneous approximation of it, trading accuracy for online availability.

\subsection{Image Association and Projection}

Each instability measurement is associated with the terrain location probed during the corresponding stance, and projected into camera images acquired before touchdown. 
For this spatial association, we rely on the exteroceptive poses computed using FAST-LIVO2~\cite{zheng2024fast}, which remain unaffected by foot slippage and thus prevent instability events from shifting the projected footholds.
The world-to-camera transformation for an image acquired at time $t$ is ${}^{\mathcal C}_{\mathcal W}\mathbf T(t) = {}^{\mathcal C}_{\mathcal B}\mathbf T\,{}^{\mathcal B}_{\mathcal W}\mathbf T(t)$, where ${}^{\mathcal C}_{\mathcal B}\mathbf T$ is the camera extrinsic calibration, obtained with the  \href{https://github.com/ethz-asl/kalibr}{Kalibr} toolbox.
The touchdown location of stance $i$ is computed from the synchronized robot pose and \ac{FK} foot position at touchdown, expressed in the world frame, and subsequently transformed into the camera frame.
Corresponding pixel coordinates are then computed with the calibrated pinhole-camera model.

We retain projections that correspond to a touchdown occurring at most \SI{10}{s} after the image acquisition, at a distance between \SI{0.5}{m} and \SI{3}{m} in front of the camera.
Each foothold is rasterized as the image projection of a \SI{6}{cm} circle, the diameter of the robot's foot, lying flat on the ground at the contact location.
When multiple projected footholds overlap, their instability scores are averaged. 

\subsection{Visual Instability Predictor}
\label{sec:metho_instability_predictor}

\begin{figure}[hb]
    \centering
    \includegraphics[width=0.48\textwidth]{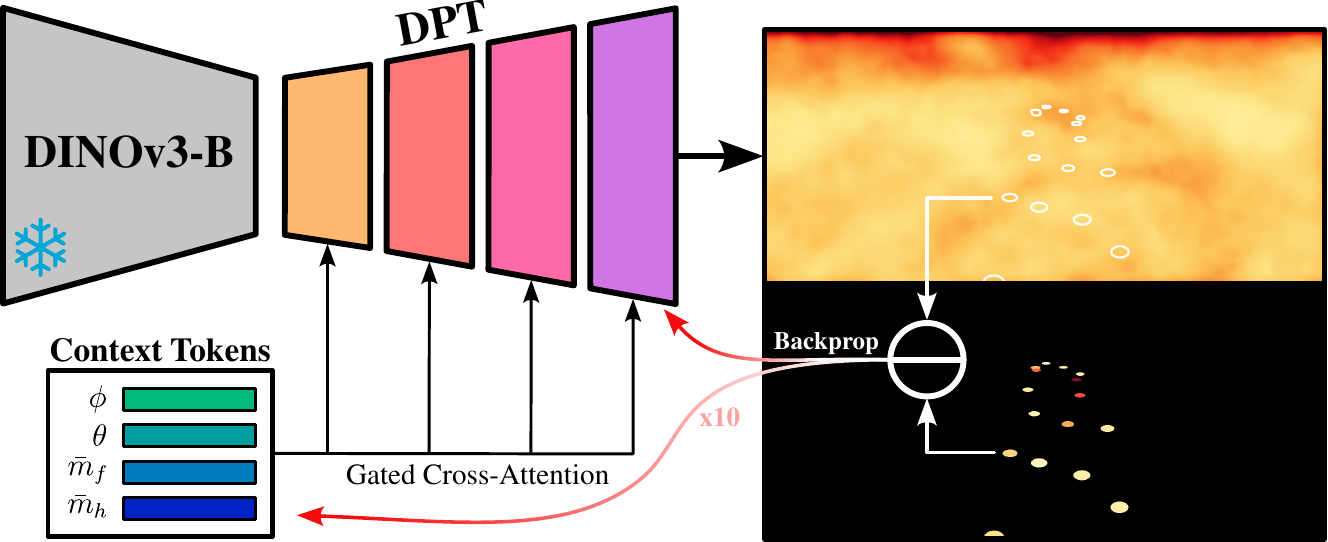}
    \caption{Visual instability predictor. A frozen DINOv3-B encoder feeds a trainable \acs{DPT} decoder that outputs a dense instability map. Camera roll and pitch ($\phi,\theta$), and recent instability history ($\bar{m}_f,\bar{m}_h$), enter as context tokens through \acs{GCA} at each decoder stage. The loss is applied only at projected footstep locations.}
    \label{fig:metho_neural_net}
\end{figure}

The projected footsteps provide sparse image-space supervision, while the objective is to predict foothold instability densely over the visible terrain.
We use a \ac{DPT}~\cite{ranftl2021vision} with a frozen DINOv3-Base encoder~\cite{simeoni2025dinov3}, as illustrated in \autoref{fig:metho_neural_net}.
The \ac{DPT} architecture suits our task because it regresses a continuous value per pixel while preserving sharp boundaries, as in monocular depth estimation.
Its decoder progressively fuses features from four intermediate layers into a single-channel, full-resolution prediction.

We supplement the prediction head with the camera roll and pitch ($\phi, \theta$) and a short instability history, averaged over preceding footsteps separately for the front ($\bar{m}_f$) and hind feet ($\bar{m}_h$).
Each context value is normalized and embedded into its own learned token. 
These tokens provide the keys and values for \ac{GCA} at each of the four DPT decoder stages, where image features act as queries, allowing each image location to weight the contextual inputs differently.
The attention output is projected to match the decoder feature dimension, then added to the decoder features through a learned scalar gate. 
The gate is initialized to zero at each stage, so conditioning initially has no effect.
We scale the learning rate of the \ac{GCA} parameters by a factor of ten, so that the gates open within the training budget.
During training, the encoder remains frozen, while the decoder and context conditioning modules are optimized.

The residual-based instability score exhibits a long-tailed distribution.
We therefore apply a logarithmic transformation to the projected labels before training.
Supervision loss is applied only to pixels covered by projected footsteps, while being ignored for all remaining pixels.
We optimize the predictor using a masked Huber loss, which reduces the influence of large prediction errors compared with a squared-error loss while retaining a quadratic penalty near zero.

\subsection{Footstep Planning}
\label{sec:footstep_planning}

We integrate the visual predictor into a simple foothold adaptation planner running onboard.
The low-level controller of ANYmal is a learned policy driven by body velocities, which offers no control over where the feet land.
We therefore replace it with the contact-tracking policy of \citet{atanassov2025gecco}, which accepts desired contact positions as input and runs at \SI{50}{Hz}.

Predictions are accumulated in a \SI{5}{m} square costmap at \SI{4}{cm} resolution, centered on the robot.
Since the camera provides no depth, the terrain is approximated by a plane fitted to the recent touchdowns of the four feet.
Each pixel of the predicted instability map is projected onto this plane, and then fused into the cell through an \ac{EMA} weighted by the inverse distance to the robot.

At each swing phase, the gait generator produces a nominal foothold $\mathbf x_0$.
The candidate set $\mathcal C$ gathers every map cell within \SI{25}{cm} of $\mathbf x_0$ that lies in the leg workspace and covers only observed cells.
We then command the foothold
\begin{equation}
    \mathbf x^\star = \operatorname*{arg\,min}_{\mathbf x\in\mathcal C}\;
    \left[\,\bar c(\mathbf x) + w\,\|\mathbf x-\mathbf x_0\|^2\,\right],
    \label{eq:foothold_selection}
\end{equation}
where $\bar c(\mathbf x)$ is the worst cost of the map over the foot contact patch centered on $\mathbf x$ and $w$ weights the deviation from the nominal foothold.

\section{Experiments and Results}
\label{sec:results}



We evaluate the approach in three stages. 
We first validate the instability score against manually annotated contacts, then assess whether it can be predicted from images alone on unseen trajectories, and finally deploy the predictor on a robot for visual footstep planning.

\subsection{Instability Score Evaluation}
\label{sec:metric_eval}

We collected a controlled dataset in three scenarios chosen to induce different forms of foothold instability, namely an ice rink partially covered with carpet, a pile of wood logs, and a slippery \SI{45}{\degree} slope.
Our ANYmal-D robot carried two GoPro cameras directed toward its feet, highlighted in \autoref{fig:intro}, continuously recording the foot-terrain interactions.
To limit labeling errors, two annotators inspected these recordings frame by frame and labeled 1716 footholds, of which 275 were considered unstable.
Every method, including ours, was tuned on this dataset, with the PCA-based scores fitted on the same data, so that each is compared at its best achievable performance.

\begin{figure}[tbp]
    \centering
    \includegraphics[width=\linewidth]{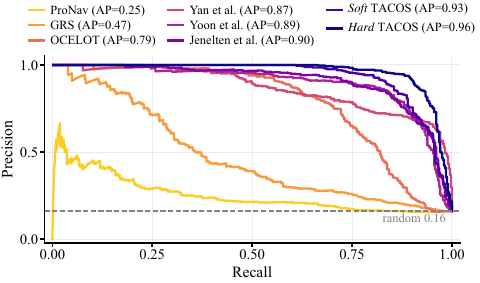}
    \caption{Precision-Recall curves for unstable-foothold classification on the human-labeled dataset, with each metric's \acf{AP} in the legend. Our \textit{hard} and \textit{soft} residuals are compared against state-of-the-art alternatives. The dashed line marks the random baseline, set by the proportion of unstable footholds in the dataset.}
    \label{fig:evaluate_metrics_pr}
\end{figure}

Both \ac{TACOS} variants achieve higher overall instability-detection \ac{AP} than all evaluated baselines.
Since unstable contacts account for only \SI{16}{\%} of the dataset, we report \ac{PR} curves, which remain informative under such class imbalance~\citep{saito2015precision}, and this proportion sets the random baseline.
As shown in \autoref{fig:evaluate_metrics_pr}, the \textit{hard} and \textit{soft} scores reach $0.96$ and $0.93$, respectively, compared with $0.90$ for the method of \citet{jenelten2019dynamic}.
The \textit{hard} variant maintains near-perfect precision up to approximately \SI{75}{\%} recall and stays above every baseline over most of the recall range.
The slight advantage of \citet{yan2024slip} at very high recall may reflect sensitivity to brief foot motion with little accumulated displacement, but comes after a larger loss of precision at lower recall.
Since these scores serve as self-supervision labels, every mislabeled foothold propagates noise into training, making precision at high recall the decisive factor.


\begin{figure}[b]
    \centering
    \includegraphics[width=\linewidth]{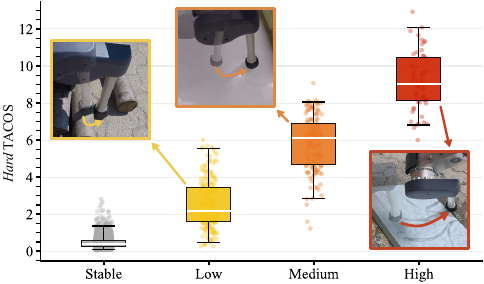}    
    \caption{Distribution of the proposed \textit{hard} \ac{TACOS} metric for stable footholds grouped by annotated instability severity. The insets show representative events recorded by the feet-facing cameras for each severity. Stable footholds concentrate at low values, and the distributions shift upward at each severity level.}
    \label{fig:evaluate_metric_severity}
\end{figure}

Beyond detection, we examine whether the residual magnitude also reflects the severity of an unstable contact.
Annotators assigned each unstable foothold a severity of low, medium or high, according to the total foot displacement observed during stance.
\autoref{fig:evaluate_metric_severity} shows the distribution of the \textit{hard} \ac{TACOS} metric for stable footholds and for the three severity levels, together with a representative event for each severity level.
Stable footholds concentrate tightly at low values, whereas the three unstable groups shift progressively upward as the annotated severity increases.
Adjacent levels still overlap, which is expected given the subjectivity of the annotations.
\autoref{tab:vus_per_run} quantifies this ordering through the \ac{VUS}, \ie{} the probability that one contact drawn from each severity level is ranked in the correct order.
Both \ac{TACOS} variants again lead on every test run and on the pooled data, \textit{hard} reaching $0.893$ against $0.700$ for \citet{yan2024slip} and $0.167$ at random.
The two variants are also complementary, as \textit{hard} dominates on ice ($0.942$), where sliding is sustained, while \textit{soft} leads on logs ($0.880$), where motion is abrupt.
The baseline ordering differs from the detection results. 
While \citet{jenelten2019dynamic} was the strongest baseline at detection, it falls behind both \citet{yan2024slip} and \citet{yoon2023invariant} at grading severity, indicating that detecting instability does not imply being able to grade it.
The two lowest scores, \ac{GRS}~\citep{wellhausen2019should} and ProNav~\citep{elnoor2024pronav}, both reduce proprioceptive features to a principal component, whose leading direction captures the dominant source of variance rather than instability itself, whereas the three strongest baselines are kinematic. 
This is consistent with our evaluation, which defines instability severity through observed foot displacement during stance.
The \ac{TACOS} metrics are kinematic as well, but read the contact residual directly from the factor graph rather than a velocity smoothed from it~\citep{yoon2023invariant}. 
The residual accumulates constraint violation over the stance, providing a continuous measure that captures both the occurrence and the severity of instability.

\begin{table}[htbp]
    \centering
    \small
    \renewcommand{\arraystretch}{1.2}
    \caption{\acf{VUS} per method, per test run and pooled over all runs. Higher is better; best per column in \textbf{bold}, second best \underline{underlined}.}
    \begin{tabular*}{\linewidth}{@{\extracolsep{\fill}}lccc|c@{}}
      \hline
      Method & Logs & Ice & Slope & Pooled \\
      \hline
      ProNav~\citep{elnoor2024pronav} & 0.341 & 0.323 & 0.312 & 0.342 \\
      GRS~\citep{wellhausen2019should} & 0.479 & 0.339 & 0.245 & 0.333 \\
      OCELOT~\citep{girgin2026ocelot} & 0.313 & 0.459 & 0.440 & 0.400 \\
      \citet{yan2024slip} & 0.693 & \underline{0.776} & 0.773 & 0.700 \\
      \citet{yoon2023invariant} & 0.797 & 0.582 & 0.777 & 0.653 \\
      \citet{jenelten2019dynamic} & 0.749 & 0.526 & 0.700 & 0.566 \\
      \textit{Soft} TACOS & \textbf{0.880} & 0.709 & \underline{0.930} & \underline{0.777} \\
      \textit{Hard} TACOS & \underline{0.838} & \textbf{0.942} & \textbf{0.939} & \textbf{0.893} \\
      \hline
    \end{tabular*}
    \label{tab:vus_per_run}
\end{table}

\subsection{Predicting Instability from Vision}

Foothold instability can be inferred from visual appearance, and the choice of supervisory signal largely determines how well.
We demonstrate this using the GrandTour dataset~\citep{tuna2026grandtour}, collected with an ANYmal-D quadruped, from which we selected 23 outdoor runs in unstructured environments.
We use the TierIV HDR camera feed, sampled every second, and discard the upper half of each image, as footholds lie below the horizon, for a total of \num{7543} images.
To limit data leakage between training and validation samples, each run is divided into contiguous temporal segments, with \SI{10}{\%} kept for validation.
Every quantitative result is reported over five cross-validation folds, which limits the influence of training stochasticity on the comparison.
Three runs are withheld entirely as the test set, namely SBB-2, KAB-2 and ALB-3, providing evaluation on trajectories not observed during training or validation.
The prediction head is trained with AdamW at a learning rate of $3 \times 10^{-5}$ and weight decay of $0.01$, under a cosine schedule with $100$ warm-up steps, using \texttt{bfloat16} precision, gradient clipping at $1.0$ and an effective batch size of $252$, for up to $100$ epochs with early stopping on validation loss.
Images are resized and randomly cropped to $512\times704$ pixels and horizontally flipped for data augmentation.

\begin{figure}[tbp]
    \centering
    \includegraphics[width=\linewidth]{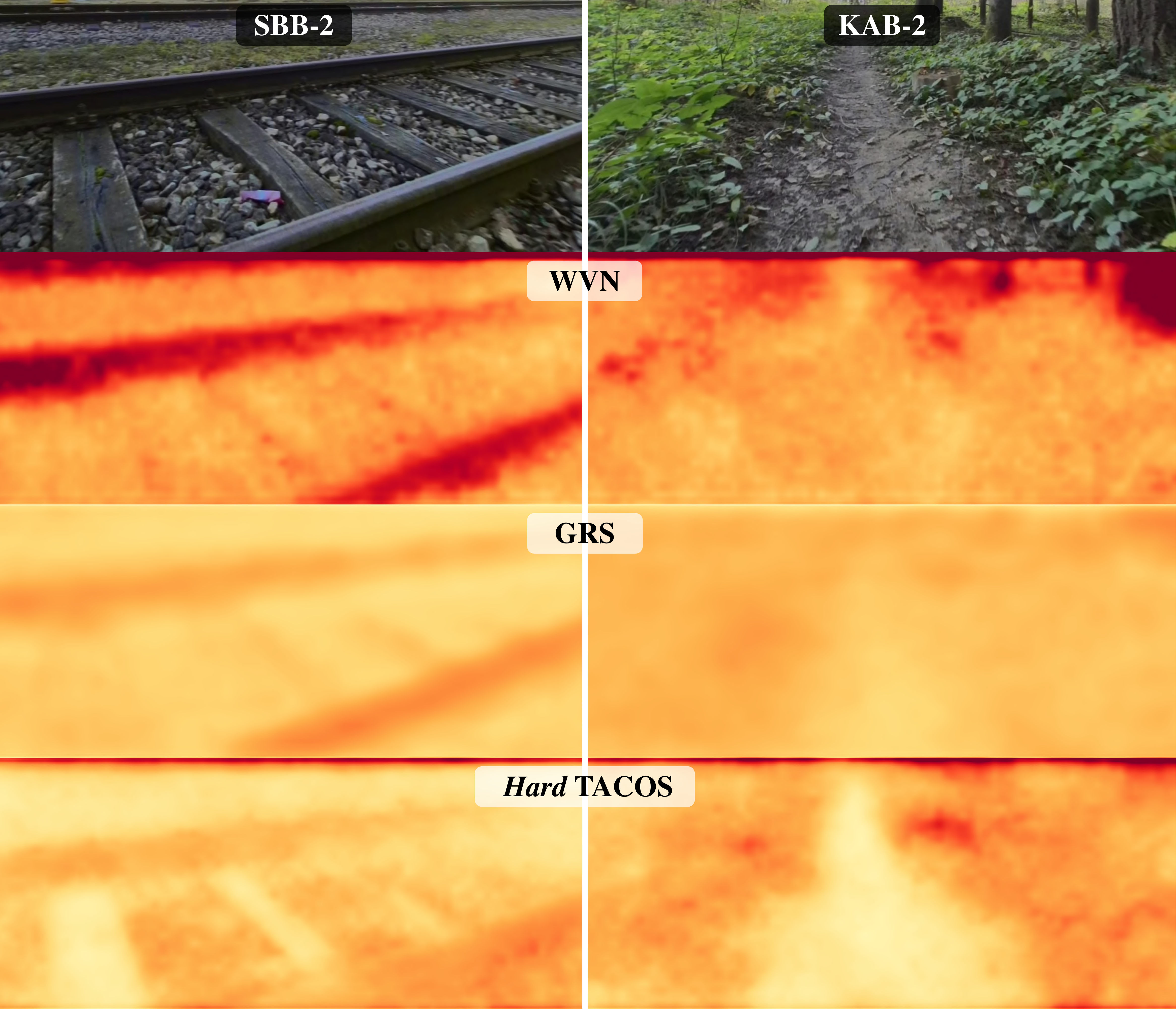}
    \caption{Network predictions on two unseen outdoor scenes for three supervisory signals, with the input image at the top. \acs{WVN} is supervised at body level, whereas \acs{GRS} and ours share the same per-footstep labels, while their metrics differ. Darker colors indicate higher predicted cost, each map being normalized between the 5th and 95th percentiles of its own training labels.}
    \label{fig:visual_qualitative}
    \vspace{-10pt}
\end{figure}

\textbf{\textit{Qualitative Evaluation --}} \autoref{fig:visual_qualitative} visually compares predictions from \ac{WVN}~\citep{mattamala2025wild}, \ac{GRS}~\citep{wellhausen2019should}, and our \textit{hard} residual on two unseen images from GrandTour.
We reproduce \ac{WVN} by projecting its traction metric onto the robot's traversed trajectory, without expanding the labels, and train it on our network.
Each costmap is normalized between the 5th and 95th percentiles of its own training labels, so that methods with different label scales remain visually comparable.
\ac{WVN} produces a coarse, high-level costmap, as expected from body-level supervision, which spans the traversed region rather than a single foot contact and labels roughly 15 times more pixels per image.
The two bottom predictors share the same per-footstep masks and differ only in the metric producing the labels, yet \ac{GRS} yields a blurry and indecisive map, whereas \textit{hard} \ac{TACOS} resolves finer terrain structure, separating the wooden railway supports from the surrounding gravel and the forest path from the adjacent vegetation.

\begin{figure}[bp]
    \centering
    \includegraphics[width=\linewidth]{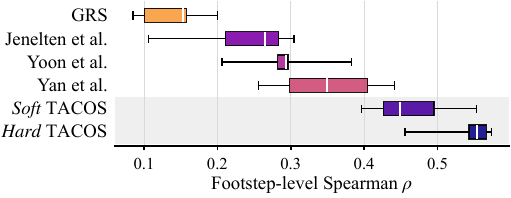}
    \caption{Footstep-level Spearman correlation between predicted and true cost labels, per label source, over five-fold cross-validation.}    \label{fig:label_source_comparison}
\end{figure}

\textbf{\textit{Quantitative Evaluation --}} Foothold selection is a ranking problem, so we compare the predicted and true instability values at every projected footstep of the test set through the Spearman rank correlation. 
Each label source carries its own units and range, so an absolute error such as the \ac{MAE} would be dominated by label scale rather than by how well the predictor recovers the ordering. 
The Spearman rank correlation instead depends only on the monotone ordering of the predictions, not on the units or range of the labels.
All predictors are trained identically on the same footstep masks, and differ only in the value assigned to each footstep.
\textit{Hard} \ac{TACOS} trains the most accurate visual predictor, as presented in \autoref{fig:label_source_comparison}, reaching a median Spearman correlation of $\rho=0.554$ over five-fold cross-validation.
\textit{Soft} \ac{TACOS} is close second at $\rho=0.449$, still ahead of every metric from \autoref{sec:metric_eval}, showing that the causal approximation remains a strong supervisory signal.
\ac{GRS}~\citep{wellhausen2019should} stays close to no correlation, as labels dominated by noise offer nothing to tie to appearance and drive the regressor toward predicting their average.
Interestingly, this ordering closely follows the one measured against human annotations in \autoref{tab:vus_per_run}, although the two evaluations share no labels, data or procedure.
The consistency of \ac{TACOS} labels therefore translates into a sharper mapping between appearance and instability, whereas noisier metrics collapse it toward the mean.

\begin{figure*}[!b]
    \centering
    \includegraphics[width=\textwidth]{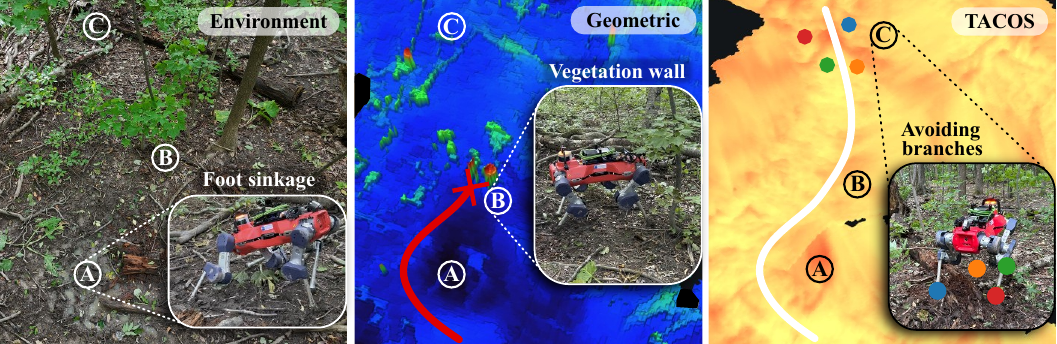}
    \caption{Real-world experiment in forest terrain. (\textbf{Left}), the geometric costmap (\textbf{Center}), and the TACOS costmap (\textbf{Right}). Corresponding locations mark \circled{A}, terrain associated with foot sinkage; \circled{B}, a vegetation barrier; and \circled{C}, fallen branches. The smaller photographs illustrate the robot’s interaction with these obstacles. Red and white curves show trajectories. The red cross marks where the geometric costmap incorrectly treats the traversable vegetation as an impassable barrier.}
    \label{fig:deployment figure}
\end{figure*}

\textbf{\textit{Ablation Study --}} Visual appearance therefore carries real signal about instability, but it cannot convey the state of the robot at the moment of image acquisition, which the network must otherwise infer from the image.
We quantify this through the context conditioning of \autoref{sec:metho_instability_predictor}, ablating the camera attitude and the recent instability history in \autoref{tab:metadata_ablation_spearman_rho_gatelr}.
Attitude alone raises the pooled correlation by \SI{4}{\%}, while the instability history alone gains \SI{25}{\%}, indicating that instability is temporally consistent, as consecutive footholds tend to probe terrain of similar nature.
Combining both adds only a marginal gain, consistent with the history already carrying most of that information.
Attitude is therefore a cheap substitute when the residual cannot be computed onboard, since it comes directly from the \ac{IMU} and still recovers a sizable part of the improvement.


\begin{table}[htbp]
    \centering
    \small
    \renewcommand{\arraystretch}{1.25}
    \setlength{\tabcolsep}{3pt}
    \caption{Ablation of context conditioning with \acs{GCA}. All configurations use images, and check marks indicate the added roll/pitch ($\phi, \theta$) and instability history ($\bar{m}_{f,h}$). Spearman correlation between predictions and \textit{hard} residual labels is reported per test run and pooled, as mean $\pm$ standard deviation over 5 folds. Higher is better, with best values in \textbf{bold} and second best \underline{underlined}.}
    \begin{tabular*}{\linewidth}{@{\extracolsep{\fill}}cc|ccc|c@{}}
      \hline
      $\phi / \theta$ & $\bar{m}_{f,h}$ & SBB-2 & KAB-2 & ALB-3 & Pooled \\
      \hline
      -- & -- &
      0.510{\scriptsize$\pm$.023} & 0.501{\scriptsize$\pm$.020} & 0.373{\scriptsize$\pm$.026} & 0.557{\scriptsize$\pm$.007} \\
      $\checkmark$ & -- &
      0.509{\scriptsize$\pm$.027} & 0.522{\scriptsize$\pm$.029} & 0.377{\scriptsize$\pm$.041} & 0.580{\scriptsize$\pm$.015} \\
      -- & $\checkmark$ &
      \underline{0.584}{\scriptsize$\pm$.018} & \underline{0.633}{\scriptsize$\pm$.016} & \underline{0.612}{\scriptsize$\pm$.018} & \underline{0.694}{\scriptsize$\pm$.007} \\
      $\checkmark$ & $\checkmark$ &
      \textbf{0.589}{\scriptsize$\pm$.022} & \textbf{0.635}{\scriptsize$\pm$.009} & \textbf{0.613}{\scriptsize$\pm$.017} & \textbf{0.694}{\scriptsize$\pm$.008} \\
      \hline
    \end{tabular*}
    \label{tab:metadata_ablation_spearman_rho_gatelr}
    \vspace{-0pt}
\end{table}

\subsection{Real-World Experiment}
\label{sec:results-real-world}

We finally deployed the trained predictor in the footstep planner of \autoref{sec:footstep_planning}, through a real-world experiment in moderate forest terrain.
The same \SI{50}{m} trajectory was executed twice, once in geometric mode and once with \ac{TACOS}.
Images were captured by an onboard Basler ace2 Pro camera, and model inference ran entirely onboard on an NVIDIA Jetson AGX Thor at \SI{10}{\hertz}.
\autoref{fig:deployment figure} shows the environment together with the built-in geometric elevation map and the \ac{TACOS} costmap built along the way.
Geometry alone perceives the mudhole at \circled{A} as a smooth surface, which leads to foot sinkage, and treats the traversable vegetation at \circled{B} as an impassable barrier, marked by the red cross.
The \ac{TACOS} costmap instead captures both, and at \circled{C} it leads the robot to step around the fallen branches rather than onto them.
While this demonstration remains limited in scope, it exemplifies the benefit of bringing terrain-aware contact supervision into footstep planning.

\section{Conclusion}
\label{sec:conclusion}


We presented TACOS, a self-supervised framework for predicting foothold instability from vision. It turns factor graph contact residuals into continuous per-footstep scores that supervise a dense visual predictor.
The \textit{hard} variant exploits the full estimation window, while the \textit{soft} variant evaluates residuals causally as measurements arrive.
Both outperform every evaluated alternative at detecting unstable contacts and at ranking their severity, improving pooled \ac{VUS} by \SI{28}{\%} over the strongest baseline.
\textit{Hard} \ac{TACOS} also trained the most accurate visual predictor, leading rank correlation by \SI{58}{\%}.
These two evaluations share no data, labels or procedure, yet place the metrics in the same order.
A closed-loop demonstration in forest terrain shows the value of bringing terrain-aware contact supervision into footstep planning where geometry alone does not capture the complexity of natural environments.

Limitations include the dependency on accurate contact detection, state estimation, and timing synchronization, which are challenging to achieve in practice.
Future works will explore incorporating depth and geometry into the system, adopting probabilistic learning approaches such as adaptive binning, and enabling continual learning over successive deployments to improve the predictor's performance over time.

\section*{Acknowledgment}

This work was supported by the Canada Research Chair on Field Robotics (CRC-2023-00396), the NSERC project CRYOTIC (ALLRP 585289-23).
The first author was further supported by NSERC CGS D under grant 600814.
Special thanks to Vassil Atanassov for sharing the contact tracking policy used for the experiments in \autoref{sec:results-real-world}.

Generative AI (Claude, Anthropic) was used to assist in generating the plotting code for the data-driven result figures in \autoref{sec:results}; the underlying data, analysis, and figure content were produced and verified by the authors.



\printbibliography

\end{document}